\documentclass[runningheads]{llncs}
\usepackage[T1]{fontenc}
\usepackage{graphicx}
\usepackage{amsmath}
\usepackage{amssymb}
\usepackage{mathtools}
\usepackage{amsfonts}
\usepackage{mathrsfs}
\usepackage{algorithm}
\usepackage{hyperref} 
\usepackage{booktabs}
\usepackage{subcaption}
\usepackage{fancyhdr}
\usepackage[noend]{algpseudocode}
\usepackage{color}

\begin{document}
\title{Learning to Optimize Capacity Planning in Semiconductor Manufacturing} 

%
%
\author{
Philipp Andelfinger\inst{1}\textsuperscript{\dag} 
\and
Jieyi Bi\inst{1}\textsuperscript{\dag}\textsuperscript{*}
\and
Qiuyu Zhu\inst{1}\textsuperscript{\dag} 
\and
Jianan Zhou\inst{1}
\and
Bo Zhang\inst{1}
\and
Fei Fei Zhang\inst{2}
\and
Chew Wye Chan\inst{2}
\and
Boon Ping Gan\inst{2}
\and
Wentong Cai\inst{1}
\and
Jie Zhang\inst{1}
}

\institute{
College of Computing and Data Science, Nanyang Technological University, Singapore 
\and
D-SIMLAB Technologies Pte Ltd, Singapore
}


\renewcommand{\thefootnote}{\fnsymbol{footnote}}
\footnotetext{\textsuperscript{\dag} Equally contributed.}
\footnotetext{\textsuperscript{*} Correspondence to Jieyi Bi <\email{jieyi001@e.ntu.edu.sg}>.}
\maketitle 
\begin{abstract}
In manufacturing, capacity planning is the process of allocating production resources in accordance with variable demand.
The current industry practice in semiconductor manufacturing typically applies heuristic rules to prioritize actions, such as future change lists that account for incoming machine and recipe dedications.
However, while offering interpretability, heuristics cannot easily account for the complex interactions along the process flow that can gradually lead to the formation of bottlenecks.
Here, we present a neural network-based model for capacity planning on the level of individual machines, trained using deep reinforcement learning.
By representing the policy using a heterogeneous graph neural network, the model directly captures the diverse relationships among machines and processing steps, allowing for proactive decision-making.
We describe several measures taken to achieve sufficient scalability to tackle the vast space of possible machine-level actions.                                                                                        
Our evaluation results cover Intel's small-scale \textsc{Minifab} model and preliminary experiments using the popular SMT2020 testbed.
In the largest tested scenario, our trained policy increases throughput and decreases cycle time by about 1.8\% each.

\keywords{Deep Reinforcement Learning \and Semiconductor Manufacturing \and Optimization \and Graph Neural Network.}
\end{abstract}
\section{Introduction}
Semiconductor manufacturing is highly complex, involving hundreds of machines and re-entrant operations~\cite{Monch2011AOperations}.
As a result, capacity planning plays a critical role in allocating resources to meet product demand and maintain target performance metrics~\cite{gopalakrishnan2018machine}.
Typical decisions in capacity planning include changes to machines and their {recipe} dedication, improving machine uptimes, and improving machine throughputs~\cite{sooleen2019}, which are captured in so-called future change lists.
In current industry practice, static and dynamic capacity planning tools rely heavily on expert knowledge to identify potential capacity problems and create appropriate future change lists~\cite{sooleen2019}.
These decisions are driven by combinations of expert knowledge and heuristics.
While heuristics rules may consider certain observable statistics of the fab operations, they cannot fully capture the dynamic relations among machines and operations.
Moreover, small variations in machine uptime and product mix can cause negligible or enormous impacts on factory performance at different periods and are difficult to consider as part of a heuristic~\cite{sooleen2019}.
Hence, heuristic-driven capacity planning is likely to leave significant opportunities for process improvements untapped.

In the present work, we explore the use of deep reinforcement learning (DRL) to train a capacity planning policy that considers the dynamically evolving fab situation and proposes suitable actions in order to improve key performance indicators such as throughput and cycle time.
In contrast to classical optimization approaches (e.g.,~\cite{gopalakrishnan2018machine}), the evolving fab state is represented by discrete-event simulations, which allows for detailed consideration of the time-varying machine and operation features, but also introduces higher complexity and significant computational demands.
A heterogeneous graph neural network (HGNN) captures the interactions of nodes and the associated operations, extending the scope of the decision-making beyond the local machine level.
We present our formulation of the Markov decision process underlying the reinforcement learning task and describe in detail our overall training framework.
An industry-grade simulation software suite forms the environment and is exercised in a parallelized setting to quickly gather a diverse set of fab state trajectories and the consequences of capacity planning decisions.
Our experiments are targeted towards understanding the scalability of our framework.
To this end, we evaluate its performance both for the small-scale \textsc{Minifab} testbed~\cite{spier1995simulation} and the larger SMT2020~\cite{kopp2020smt2020}, showing substantial improvements in performance indicators in both cases.

\section{Related Work}

Traditionally, capacity planning approaches from the literature take the form of mathematical models that are solved heuristically to tackle the problem scales of real-world fabs~\cite{repec:spr:sprbok:978-1-0716-0354-3}.
In contrast, data-driven approaches using observations from simulations or real-world fabs can more closely represent and exploit the complex relationships among fab states, actions, and performance indicators~\cite{wang2018data,meidan2011cycle}.

To exploit learned relationships for planning under unpredictable situations such as unplanned downtimes, digital twin architectures that allow for rapid decision-making employ learned models, often based on reinforcement learning (RL).
A comprehensive survey covering works on capacity planning using RL until 2021 is provided by Kulmer et al.~\cite{kulmer2022medium}.

Waschneck et al.~employed cooperative Deep Q-Learning (DQN) agents for scheduling in a semiconductor fab comprised of four machine groups~\cite{waschneck2018optimization}.
Agents representing machine groups were trained one by one to carry out local scheduling decisions, while observing actions of other agents.

Lee et al.~applied a DQN to schedule lots in a manufacturing line covering 60 machines and ten wafer types~\cite{lee2022deep}.
Here, the reward measures the adherence of the chosen actions with a pre-determined production plan that prescribes the quantities to be produced, generated by solving a linear program.

Capacity planning for a cement factory was studied by Danket et al.~\cite{danket2024mixture}.
As in our work, they rely on deep RL via Proximal Policy Optimization (PPO).
However, in place of our HGNN, deep feed-forward networks represent the actor and critic.
Taking a mixture-of-experts approach, four agents are trained on different demand patterns.
During inference, a gating agent aims to select the best-suited expert for observed emerging demand patterns.

Kuo et al.~presented a DQN-based digital twin framework for semiconductor manufacturing, focused on handling demand uncertainty~\cite{kuo2025deep}.
In contrast to other works, the learned policy does not propose actions directly, but selects from a pool of traditional mathematical models.

Tassel et al.~trained a DRL-based dispatching policy using a feed-forward network with double-headed self-attention~\cite{tassel2023semiconductor}.
At each step, the policy assigns scores to define an order on the waiting lots, which are represented by a combination of lot-level features and features describing applicable machine families.

Recently, HGNNs have been employed in DRL using PPO with great success to solve flexible job-shop scheduling problems~\cite{peng2025heterogeneous,tang2024solving,song2022flexible}.
By capturing interactions across machines and operations, richer policies can be learned.
However, to our knowledge, HGNNs have not previously been applied to simulation testbeds for semiconductor manufacturing.

Unlike prior works, we focus on actions on the planning rather than operation level. Our experiments include capacity planning for the SMT2020 testbed, which covers more than 1,300 machines and over 4,000 operations.

\section{Preliminaries} 
\subsection{Problem Definition and MDP formulation}
\label{sec:preliminaries:subsec:problem_definition}

Capacity planning (CP) aims to dynamically allocate production resources to accelerate the manufacturing process. In semiconductor manufacturing, a CP instance consists of a product set $\mathcal{P}$ (different models of lots of wafers), a machine set $\mathcal{M}$ and an operation set $\mathcal{O}$. In particular, the operation types include lithography, diffusion, dry etching, wet etching, implantation, etc. 
Each product $p \in \mathcal{P}$ consists of $n_p$ operation steps $\{O_{p, j}\}_{j=1}^{n_p} \subseteq \mathcal{O}$ required to be processed subject to precedence constraints, i.e., $O_{p, 1} \rightarrow O_{p, 2} \rightarrow \cdots \rightarrow O_{p, n_p}$. Each operation $O_{p, j}$ can only be processed by any of the compatible machines $\{m_1, m_2, ..., m_k\} = M_{p, j} \subseteq \mathcal{M}$, where $m_k$ represents a single machine. We consider three types of decision variables in this paper: (1) \textit{dedication}, assigning machines to operations $d^{+}(m_k, O_{p, j}): \mathcal{M} \rightarrow\mathcal{O}$ or removing the assignment $d^{-}(m_k, O_{p, j}): \mathcal{M} \rightarrow\mathcal{O}$; 
(2) \textit{uptime} $u(m_k)$, increasing a machine's percentage of uptime by 3\% (but not exceeding 100\%); and (3) \textit{efficiency} $r(m_k)$, reducing a machine's per-lot processing time to 90\% of its original value. These decision types follow those in~\cite{sooleen2019}, while the specific parameter values are defined in this work. The objective of a CP solution is to maximize the total production output $N=\sum p$ and minimize the average cycle time $T=\frac{1}{N}\sum_{p=1}^N T_p$, subject to a limited number of changes to the decision variables. 

We formulate the long-horizon capacity planning as an MDP $(\mathcal{S}, \mathcal{A}, \mathcal{T}, \mathcal{R}, \gamma < 1)$. At each step $t$, the \textbf{state} $s_t$ describes the factory status, and the \textbf{action} $a_t$ consists of $\sigma$ changes to decision variables. In the \textsc{Minifab} setting, $a_t = \{u(m_i)\}_{i=1}^{\sigma_{u}} \cup \{r(m_i)\}_{i=1}^{\sigma_{r}}$ with $\sigma_u = \sigma_r = 1$. In contrast, the \textsc{SMT2020} setting uses the full action set: $a_t = \{u(m_i)\}_{i=1}^{\sigma_u} \cup \{r(m_i)\}_{i=1}^{\sigma_r} \cup \{d^-(m_i, O_{p,j})\}_{(i,p,j) \in \mathcal{I}_{d^-}} \cup \{d^+(m_i, O_{p,j})\}_{(i,p,j) \in \mathcal{I}_{d^+}}$, with $|\mathcal{I}_{d^-}| = \sigma_{d^-}$, $|\mathcal{I}_{d^+}| = \sigma_{d^+}$ and $\sigma_u = \sigma_r = \sigma_{d^-} = \sigma_{d^+} = 5$. The action space in \textsc{SMT2020} is substantially larger, as $u(m_i)$ and $r(m_i)$ are defined over $|\mathcal{M}|$ machines, while $d^+$ and $d^-$ span $|\mathcal{M}| \times |\mathcal{O}|$ combinations, introducing significant complexity. The \textbf{state transition} function $\mathcal{T}$ is deterministic and updates $s_t$ to $s_{t+1}$ based on the simulation behavior. At each decision step $t$, we evaluate the simulation both with and without applying the selected actions. This allows us to better distinguish the effects of actions from the simulations' stochasticity by defining the \textbf{reward} as $r_t = \text{KPI}_1 - \text{KPI}_0$, where $\text{KPI}_1$ and $\text{KPI}_0$ denote the performance metric with and without applying the selected actions. In this work, we adopt the fab-level daily going rate as the reward KPI, which is defined as 
$
DGR = \sum_{i=1}^{I} \frac{r_i }{L_i} \sum_{l=1}^{L_i} DGR_{i, l}
$, where $DGR_{i, l}$ is the daily output of operation $l$ for product $i$, $L_i$ is the number of operations of product $i$, and $r_i$ is the WIP ratio of product $i$.

\subsection{Simulation Environment}
\label{sec:preliminaries:subsec:simulation_environments}

We consider two common semiconductor manufacturing models, \textsc{SMT2020} and \textsc{Minifab}, whose key properties are summarized in Table~\ref{tab:stat}.

\begin{table}[htbp]
\vspace{-15pt}
\caption{Statistics of the \textsc{SMT2020} and \textsc{Minifab} testbeds.}
    \setlength{\tabcolsep}{6pt}
    \scriptsize
    \centering
    \begin{tabular}{cccccc}
    \toprule
       Model  & ~|$\mathcal{P}$| & ~|$\mathcal{M}$|& ~|$\mathcal{O}$|&  |$\mathcal{O}_p$| \\
       \midrule
       \textsc{SMT2020}  &  10 & 1,314 & 4,014 & 242 to 583 \\
       \textsc{Minifab} &   3 & 5 & 18 & 6\\
    \bottomrule
    \end{tabular}
    \vspace{10pt}
    \label{tab:stat}
\vspace{-25pt}
\end{table}

Intel’s \textsc{Minifab}~\cite{spier1995simulation} models a small fab with five machines (two for diffusion, two for ion implantation, one for lithography) and three products with six operations each, incorporating typical wafer fab attributes such as batch processing, reentrant operations, and operation-dependent setups~\cite{hassoun2019new,pfund2008multi,driessel2012integrated}; while detailed enough for controlled studies, it is far smaller than real-world fabs. In contrast, the SMT2020 testbed~\cite{kopp2020smt2020} represents 1,314 machines and up to 583 processing steps per product, with two datasets, ``high-volume, low-mix'' (HV/LM) and ``low-volume, high-mix'' (LV/HM), commonly used to evaluate dispatching and scheduling approaches~\cite{monch2007simulation,kovacs2022customizable,tassel2023semiconductor}. We use the more diverse LV/HM dataset, increasing wafer arrivals by 25\% to accentuate bottlenecks and test scalability. Both models are simulated in D-SIMCON~\cite{dsimlab2023}, a discrete-event engine with Python interface for optimization and machine learning~\cite{Schulz2022,Chan2023,stockermann2025}, which captures process flows, dedication, wafer starts, downtimes~\cite{seidel2017}.
D-SIMCON supports state copying during policy rollouts~\cite{Bosse2024}, which allows us to periodically realign a no-action baseline trajectory for variance reduction.

\section{Methodology} 

To address the capacity planning MDP formulated in Section~\ref{sec:preliminaries:subsec:problem_definition}, we train a policy $\pi_\theta$ parameterized by an encoder-decoder network. A heterogeneous graph encodes machine-operation and operation-operation interactions; the encoder generates node embeddings, which the attention-based decoder uses to compute action probabilities. The policy is trained via $n$-step PPO.


\subsection{Heterogeneous Graph Construction}

To model the complex interactions between machines and operations in a production pipeline, we construct a heterogeneous graph $\mathcal{H}$ with two types of nodes: $\mathcal{M} = \{\mathcal{M} _1, \mathcal{M} _2, \dots\}$ and operation nodes $\mathcal{O} = \{\mathcal{O}_{11}, \mathcal{O}_{12}, \dots\}$. 
Different types of edges capture key relationships in the production process: operation-operation edges reflect logical dependencies between sequential operations, while operation-machine edges denote assignment relations.
Each node and edge in the graph is associated with a feature vector extracted from manufacturing logs and configuration metadata:
\begin{itemize}
  \item \textbf{Machine node features:} $m_k \in \mathbb{R}^{d_1}$ encode machine attributes, including min/max batch sizes, waiting lot number, completed lots and wafers, average cycle time, queue time, and processing time, as well as utilization-related metrics such as productive time, down time, idle time, setup time, interval time, dispatch queue length, and end-of-period work-in-progress (WIP) lots and wafers.
  \item \textbf{Operation node features:} $o_{ij} \in \mathbb{R}^{d_2}$ capture operation-specific properties, including completed wafers, completed lots, completed wafer layers, completed main-process wafers (excluding optional steps), average WIP lots, average out-of-process wafers, average cycle time, average remaining due time, average queue time, average processing time, average remaining processing time, average operation-specific remaining due time, average fab cycle time, daily going rate, and dynamic cycle time.
  \item \textbf{Operation-Machine edge features:} $\varepsilon_{ij,k} \in \mathbb{R}^{d_3}$ represent contextual assignment data (i.e., processing time, setup cost).
  \item \textbf{Operation-Operation edge features:} $\varepsilon_{ij,i(j+1)}\in \mathbb{R}^{d_4}$ include dependency-specific information (i.e., process flow).  
\end{itemize}


\subsection{Model Architecture}

\begin{figure}[t]
\centering\includegraphics[width=0.9\textwidth] {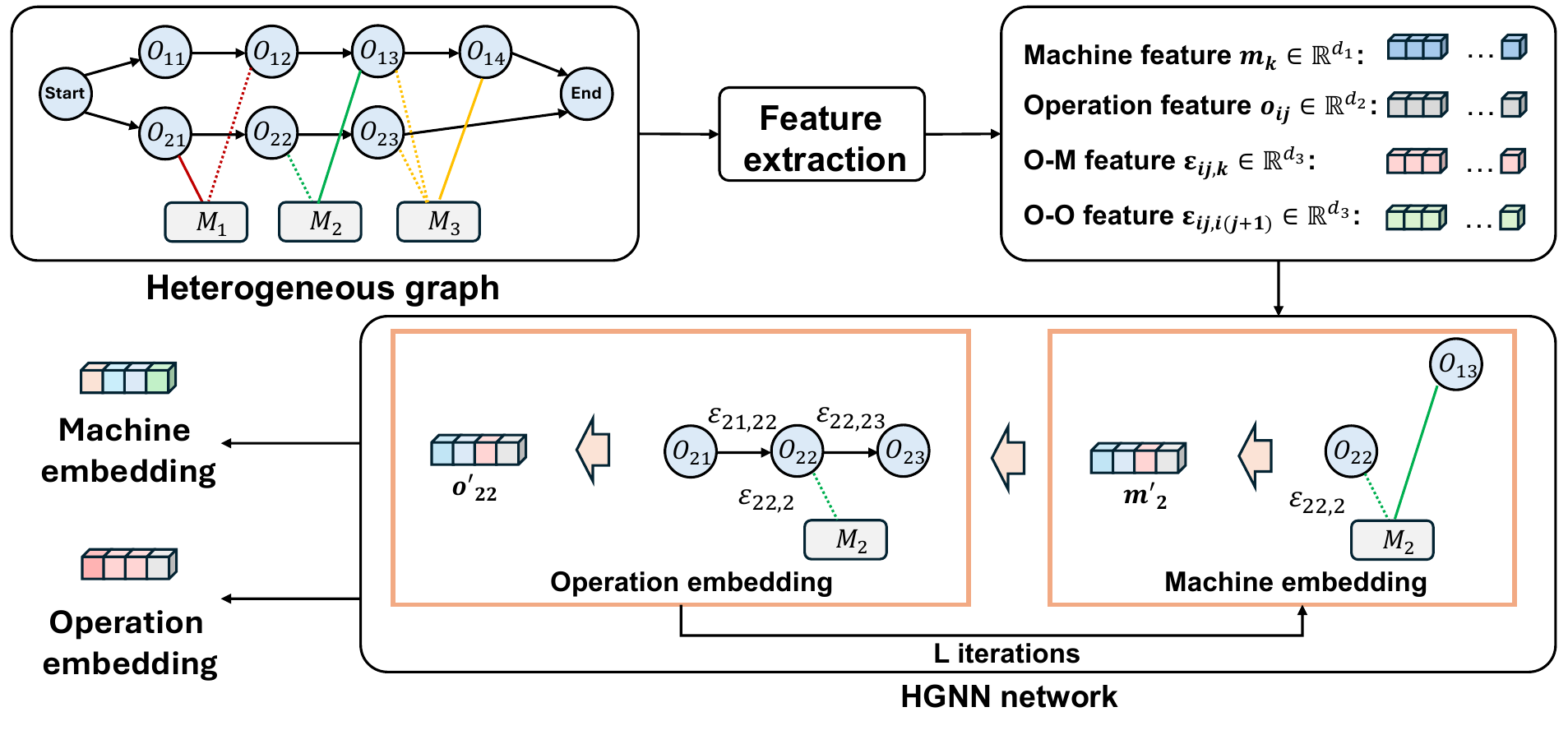}
  \caption{Overview of feature extraction and HGNN network.}  
  \label{fig:hgnn}
  \vspace{-0.4cm}
\end{figure}

Our model learns meaningful embeddings for both machine and operation nodes using an edge-aware heterogeneous graph neural network (HGNN), following \cite{song2022flexible}. As illustrated in Fig.~\ref{fig:hgnn}, the architecture consists of two main components. 
\begin{itemize}
    \item \textbf{Machine Embedding (Edge-aware Attention)}: Each machine node aggregates information from its associated operation nodes using an attention mechanism that accounts for both node features and edge attributes. This allows the network to capture the varying importance of different operations assigned to the same machine.
    \item \textbf{Operation Embedding (MLP Aggregation)}: Each operation node updates its embedding by integrating information from its immediate neighbors in the process flow (i.e., preceding and succeeding operations) as well as the embedding of the assigned machine. This is implemented using a multi-layer perceptron (MLP) to flexibly combine diverse sources of information.
\end{itemize}

These two components are applied iteratively over $L$ message-passing layers to capture complex, higher-order dependencies between operations and machines. At each layer, node representations are updated based on their local neighborhoods. The final node embeddings are computed by averaging the representations across all layers:
$m_i^{\text{final}} = \frac{1}{L} \sum_{\ell=1}^{L} m_i^{(\ell)}$, and 
$o_{ij}^{\text{final}} = \frac{1}{L} \sum_{\ell=1}^{L} o_{ij}^{(\ell)}$,
where $m_i^{(\ell)}$ and $o_{ij}^{(\ell)}$ denote the embeddings of machine node $\mathcal{M}_i$ and operation node $\mathcal{O}_{ij}$ at the $\ell$-th layer, respectively.

\vspace{-0.4cm}
\subsubsection*{Attention-based Decoder.}
Final machine and operation embeddings are passed to a lightweight decoder to compute action probabilities. For machine-level uptime and efficiency, separate MLPs with sigmoid activations are applied to machine embeddings, yielding independent one-dimensional probabilities for each machine: $\pi_i^u = \text{Sigmoid}\left(\text{MLP}_u\left(e_i^{\text{final}}\right)\right)$ and $\pi_i^r = \text{Sigmoid}\left(\text{MLP}_r\left(e_i^{\text{final}}\right)\right)$.
For operation-machine dedication actions, we use an attention mechanism to model pairwise interactions between machine and operation embeddings. These interactions are then passed through a scaled tanh and sigmoid to produce normalized probabilities: $\pi_{ij}^{d^+} = \text{Sigmoid}\left(-C \cdot \text{Tanh}\left(\frac{e_i^{\text{final}} \cdot r_{ij}^{\text{final}}}{\sqrt{d}} \right)\right)$,
where $C$ is a scaling factor that encourages exploration; following~\cite{bi2024learning}, we set $C = 10$. The probability of removing a dedication is defined as the complement of the probability of adding one, i.e., $\pi_{ij}^{d^-} = 1 - \pi_{ij}^{d^+}$.

\subsubsection*{Critic Network.} 
The critic network, parameterized by $\phi$, shares the HGNN encoder with the policy network. To estimate the value of the overall factory state, it aggregates both machine and operation embeddings via separate mean pooling, which are concatenated to form the graph state embedding $h_t \in \mathbb{R}^{2d}$:
$h_t = \left[\frac{1}{|\mathcal{M}|}\sum_{m_i\in\mathcal{M}} e_i^{\text{final}}~~\vert \vert ~~\frac{1}{|\mathcal{O}|}\sum_{O_{p,j} \in \mathcal{O}} r_{ij}^{\text{final}} \right]$.
Finally, $h_t$ is projected to a scalar state value via a linear layer.

\subsection{Training Paradigm} 

\begin{figure}[t]
    \centering
    \includegraphics[width=\linewidth]{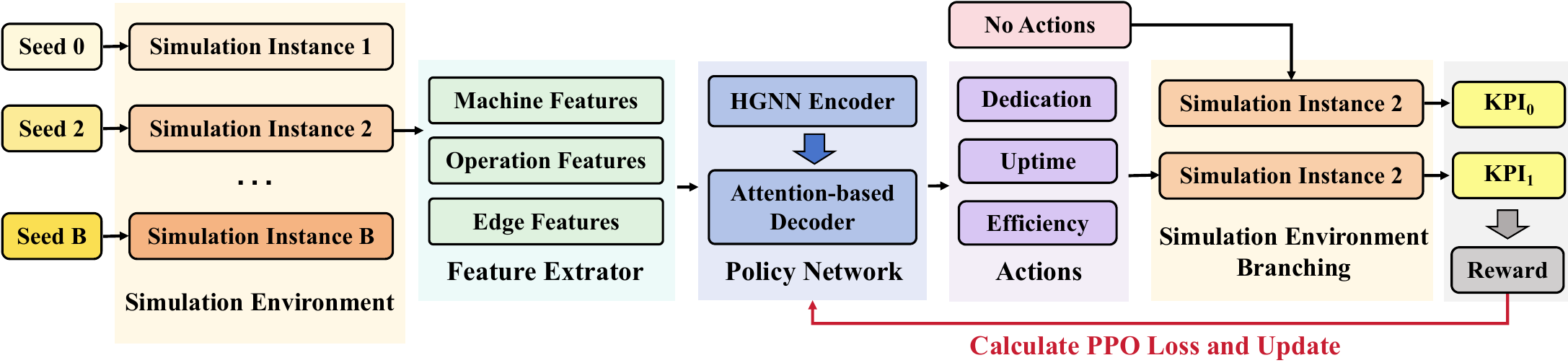}
    \caption{Overview of the training framework.}
    \label{fig:overview}
     \vspace{-5pt}
\end{figure}

As detailed in Algorithm \ref{alg}, we train the policy $\pi_{\theta}$ via the $n$-step PPO algorithm in an online setting, where our agent interacts with the simulation environment during training. As shown in Figure~\ref{fig:overview}, multiple simulation instances are executed in parallel to enable batch training. During training, the parameters of the policy network $\theta$ and critic network $\phi$ are updated accordingly.

\begin{algorithm}[t]
\caption{$n$-step Training with PPO}
\begin{algorithmic}[1]
\Require Initial \textbf{policy network} parameters $\theta$; initial \textbf{critic network} parameters $\phi$; clipping threshold $\epsilon$; learning rate $\eta_\theta$, $\eta_\phi$
\For{epoch $= 1$ to $E$}
    \For{$b = 1$ to $B$ \textbf{in parallel}}
        \State Generate environment instance $\mathcal{D}_b$ with unique random seed
        \State Set initial state $s_{b,0}$
    \EndFor
    \State $t \gets 0$
    \While{$t < T_{\text{train}}$}
        \For{$b = 1$ to $B$ \textbf{in parallel}}
            \State Collect experiences $\{(s_{b,t'}, a_{b,t'}, r_{b,t'})\}_{t'=t}^{t+n}$ by running policy $\pi_\theta$ for $n$ time
            \Statex\hspace{7cm}steps, where $a_{b,t'} \sim \pi_\theta(a_{b,t'}|s_{b,t'})$
        \EndFor
        \State $t \gets t + n$, $\pi_{\text{old}} \gets \pi_\theta$, $v_{\text{old}} \gets v_\phi$
        \For{$k = 1$ to $K$}
            \State $\hat{R}_t \gets v_\phi(s_t)$
            \For{$t' \in \{t-1, \dots, t-n\}$}
                \State $\hat{R}_{t'} \gets r_{t'} + \gamma \hat{R}_{t'+1}$
            \EndFor
            \State Normalize $\hat{R}_{t'} $ by Z-score
            \State $\hat{A}_{t'} \gets \hat{R}_{t'} {- v_\phi(s_{t'})}$
            \State Compute PPO loss $\mathcal{L}^{\text{PPO}}_\theta$ and Critic loss $\mathcal{L}^\text{Critic}_\phi$ as in \cite{schulman2017proximal}
            \State $\theta \gets \theta + \eta_\theta \nabla \mathcal{L}^\text{PPO}_\theta$
            \State $\phi \gets \phi - \eta_\phi \nabla \mathcal{L}^\text{Critic}_\phi$ 
        \EndFor
    \EndWhile
\EndFor
\end{algorithmic}
\label{alg}
\end{algorithm}

\section{Experiments}

We evaluate our method on two common simulation-based testbeds (cf.~Section~\ref{sec:preliminaries:subsec:simulation_environments}): Intel’s small-scale \textsc{Minifab} and \textsc{SMT2020}.

We train using Adam with learning rates $\eta_\theta = 3 \times 10^{-4}$ (policy) and $\eta_\phi = 10^{-4}$ (critic). PPO is configured with $n=5$, $K=20$, clipping parameter $\epsilon = 0.2$, and discount factor $\gamma = 0.99$.
We use a batch size of $B=16$.
In \textsc{Minifab}, actions are taken daily over $T_{\text{train}}=5$ simulation days; in \textsc{SMT2020}, actions are taken weekly over $T_{\text{train}}=25$ weeks. 
We set the HGNN's layer count to $L = 1$ for \textsc{Minifab}, to $L = 2$ for \textsc{SMT2020}, and the hidden dimension to $d = 64$.
As baselines, our results include three strategies: taking no actions, taking random actions, and a heuristic that samples target machines and dedications based on their work-in-progress levels.

\subsection{Results on \textsc{Minifab}} 

We first assess the HGNN's sensitivity to its hyperparameters by varying the hidden dimension $d \in {32, 64, 128, 256}$ and the layer count $L \in {1, 2}$.
Each strategy is tested across 16 simulation instances using the same sequence of random seeds to ensure fairness.
We evaluate three fab performance indicators: completed lots, average cycle time and daily going rate. 
As shown in Table~\ref{tab:hyperparameter}, the best results are obtained with $d = 64$ and $L = 1$, showing that larger dimensions or deeper networks may lead to overfitting or increased optimization difficulty.

\begin{table}[t]
\caption{Impact of HGNN hyper-parameters on performance in \textsc{Minifab}.}
\scriptsize
\setlength{\tabcolsep}{6pt}
\centering
\begin{tabular}{lccc}
\hline
\textbf{Hyperparameter} & \textbf{Completed Lots} $\uparrow$ & \textbf{Cycle Time} $\downarrow$ & \textbf{Daily Going Rate} $\uparrow$ \\
\hline
$d = 32$       & 55.06 & 1.10 & 265.73 \\
$d = 64$       & \textbf{55.50} & \textbf{1.08} & \textbf{266.56} \\
$d = 128$      & 54.69 & 1.09 & 265.26 \\
$d = 256$      & 54.69 & 1.10 & 266.26 \\
\midrule
$L = 1$        & \textbf{55.50} & \textbf{1.08} & \textbf{266.56} \\
$L = 2$        & 55.06 & 1.11 & 266.09 \\
\hline
\vspace{1pt}
\end{tabular}
\label{tab:hyperparameter}
\vspace{-10pt}
\end{table}

\begin{table}[t]
\caption{Comparison of fab performance metrics across strategies.}
\scriptsize
\setlength{\tabcolsep}{9.5pt}
\centering
\begin{tabular}{lccc}
\hline
\textbf{Strategy} & \textbf{Completed Lots} $\uparrow$ & \textbf{Cycle Time} $\downarrow$ & \textbf{Daily Going Rate} $\uparrow$ \\
\hline
GNN        & \textbf{55.50} & \textbf{1.08} & \textbf{266.56} \\
Heuristic  & 54.63          & 1.11         & 265.42 \\
Random     & 54.63          & 1.11         & 265.00 \\
No Action  & 48.75          & 1.27         & 251.09 \\
\hline
\vspace{1pt}
\end{tabular}
\label{tab:dgr_results}
\vspace{-0.4cm}
\end{table}

\begin{table}[t]
\caption{Improvement of GNN over the baseline strategies in \textsc{\textsc{Minifab}}.}
\scriptsize
\setlength{\tabcolsep}{8pt}
\centering
\begin{tabular}{lccc}
\hline
\textbf{GNN vs.} & \textbf{Completed Lots} & \textbf{Cycle Time} & \textbf{Daily Going Rate} \\
\hline
Heuristic & 1.59\% (+0.87) & 2.70\% (-0.03 days) & 0.43\% (+1.14) \\
Random    & 1.59\% (+0.87) & 2.70\% (-0.03 days) & 0.59\% (+1.56) \\
No Action & {13.85\% (+6.75)} & {14.96\% (-0.19 days)} & {6.16\% (+15.47)} \\
\hline
\end{tabular}
\label{tab:relative_improvement}
\vspace{-18pt}
\end{table}

We evaluate the effectiveness of our HGNN in the \textsc{Minifab} environment by comparing it against our baselines. 
As shown in Table~\ref{tab:relative_improvement}, our method consistently outperforms all baselines: it achieves a 1.59\% increase in completed lots and a 2.70\% reduction in cycle time compared to the heuristic and random decisions. Against the no-action baseline, the improvements are more pronounced, with 13.85\% higher throughput and 14.96\% lower cycle time.


\subsection{Results on \textsc{SMT2020}} 

\begin{figure}[t!]
  \centering
  \begin{subfigure}[b]{0.32\textwidth}
    \includegraphics[width=\textwidth] {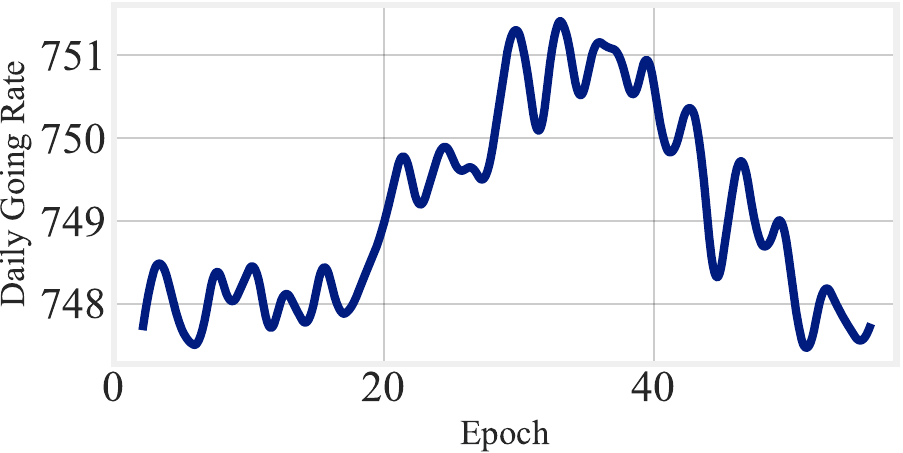}
  \end{subfigure}
  \hfill
  \begin{subfigure}[b]{0.32\textwidth}
    \includegraphics[width=\textwidth]{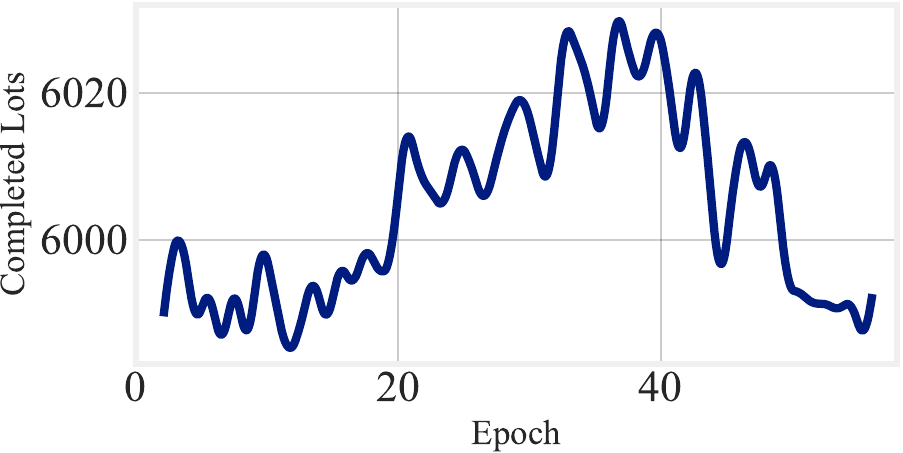}
  \end{subfigure}
  \hfill
  \begin{subfigure}[b]{0.32\textwidth}
    \includegraphics[width=\textwidth]{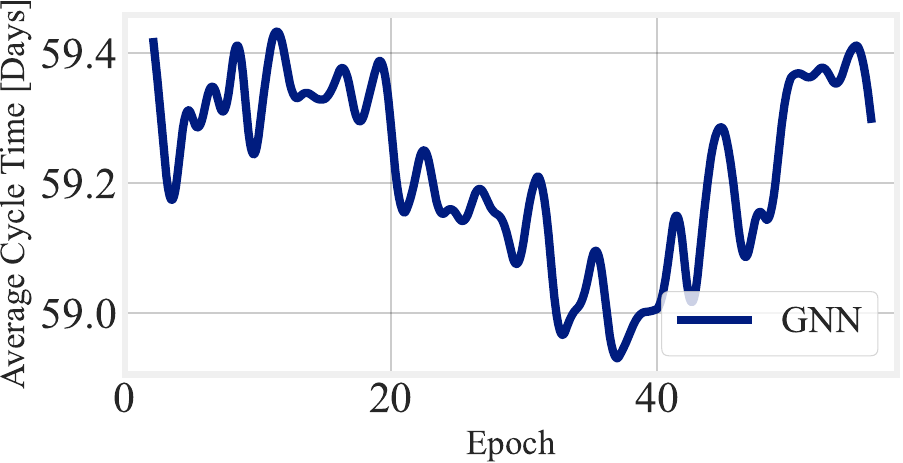}
  \end{subfigure}
  \caption{Fab-level KPIs during training in SMT2020.}
  \vspace{-0.2cm}
  \label{fig:smt2020_full_train}
  
\end{figure}

\begin{figure}[t!]
  \centering
  \begin{subfigure}[b]{0.32\textwidth}
    \includegraphics[width=\textwidth]{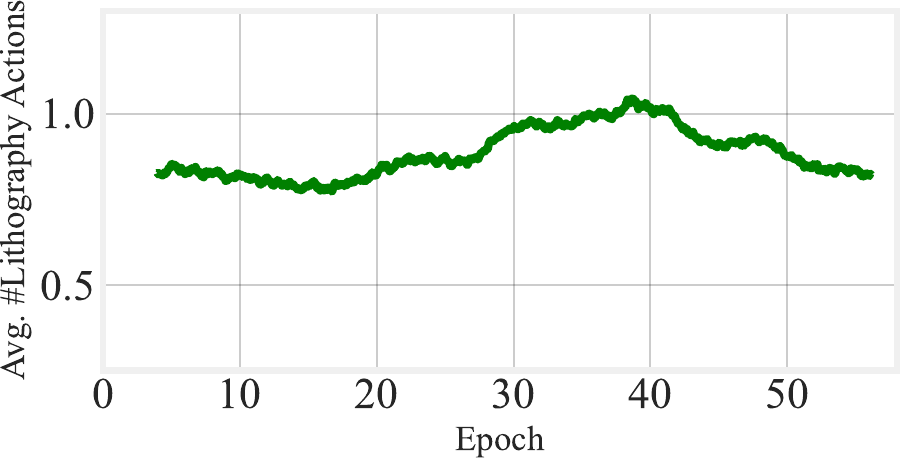}
  \end{subfigure}
  \hfill
  \begin{subfigure}[b]{0.32\textwidth}
    \includegraphics[width=\textwidth]{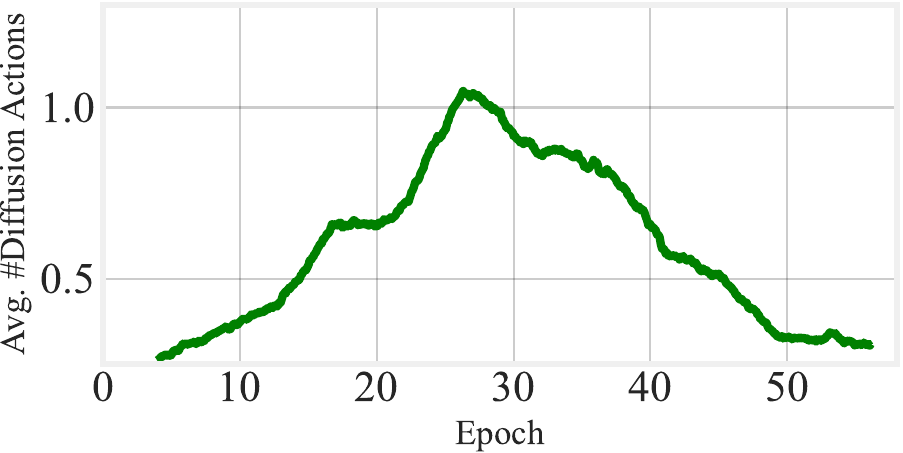}
  \end{subfigure}
  \hfill
  \begin{subfigure}[b]{0.32\textwidth}
    \includegraphics[width=\textwidth]{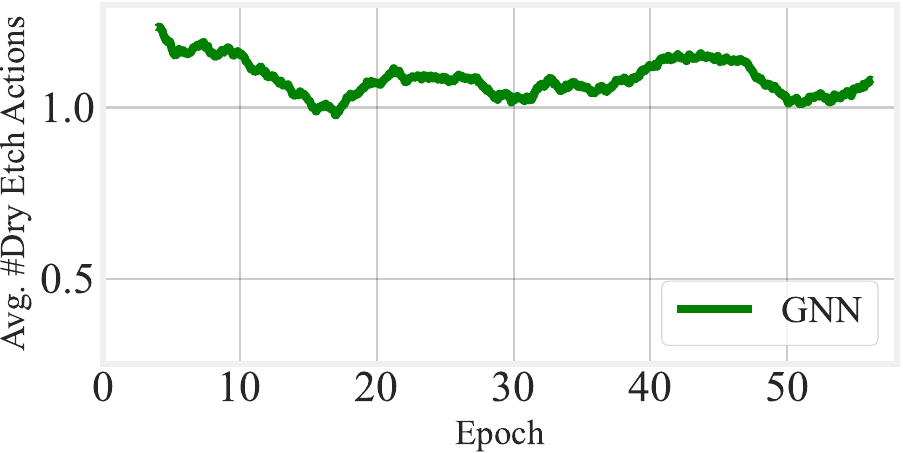}
  \end{subfigure}
    \caption{Frequency of efficiency actions on different machine families in SMT2020.}
    \vspace{-0.2cm}
  \label{fig:smt2020_full_train_actions}
  
\end{figure}

We now turn to our SMT2020 scenario, which involves a substantially larger state space and spans a duration of 25 weeks per episode.
Figure~\ref{fig:smt2020_full_train} shows the fab-level KPIs over the course of the training.
The main observation is a substantial improvement in the KPIs over the first 30 to 40 epochs, with a subsequent decline towards the initial values.
We also recorded the concrete efficiency actions taken with respect to machines of different families.
Figure~\ref{fig:smt2020_full_train_actions} shows that the policy increasingly targets lithography and diffusion before returning to the initial frequencies.
Most of the frequencies for other families such as dry etching slightly decreased, but with less clear trends over the course of the training.
At epoch 40, we evaluated against the baseline strategies. Table~\ref{tab:smt2020_improvements} shows that, in line with the training curves, our policy outperforms the other strategies substantially, with increases of 1.87\% and 1.74\% in throughput and daily going rate, and a reduction in 1.8\% in average cycle time compared to the heuristic.

\begin{table}[t]
\caption{Performance comparison across different methods in \textsc{SMT2020}.}
\scriptsize
\setlength{\tabcolsep}{8.5pt}
\centering
\begin{tabular}{lccc}
\hline
\textbf{Strategy} & \textbf{Completed Lots} $\uparrow$ & \textbf{Cycle Time} $\downarrow$ & \textbf{Daily Going Rate} $\uparrow$ \\
\hline
GNN       & \textbf{6019.75} & \textbf{59.100} & \textbf{749.79} \\
Heuristic & 5909.44 & 60.181 & 736.99 \\
Random    & 5904.13 & 60.211 & 736.65 \\
No Action & 5872.94 & 60.507 & 732.58 \\
\hline
\vspace{1pt}
\end{tabular}
\label{tab:smt2020_results}
\vspace{-25pt}
\end{table}

\begin{table}[t!]
\caption{Improvement of GNN over the baseline strategies in \textsc{SMT2020}.}
\scriptsize
\setlength{\tabcolsep}{8.5pt}
\centering
\begin{tabular}{lccc}
\hline
\textbf{\textbf{GNN vs.}} & \textbf{Completed Lots} & \textbf{Cycle Time} & \textbf{Daily Going Rate} \\
\hline
Heuristic & {1.87\% (+110.3)} & {1.80\% (-1.08 days)} & {1.74\% (+12.80)} \\
Random    & 1.96\% (+115.6) & 1.85\% (-1.11 days) & 1.78\% (+13.14) \\
No Action  & 2.50\% (+146.8) & 2.33\% (-1.41 days) & 2.35\% (+17.21) \\
\hline
\end{tabular}
\label{tab:smt2020_improvements}
\end{table}

\begin{figure}[t!]
  \centering
  \begin{subfigure}[b]{0.32\textwidth}
    \includegraphics[width=\textwidth]{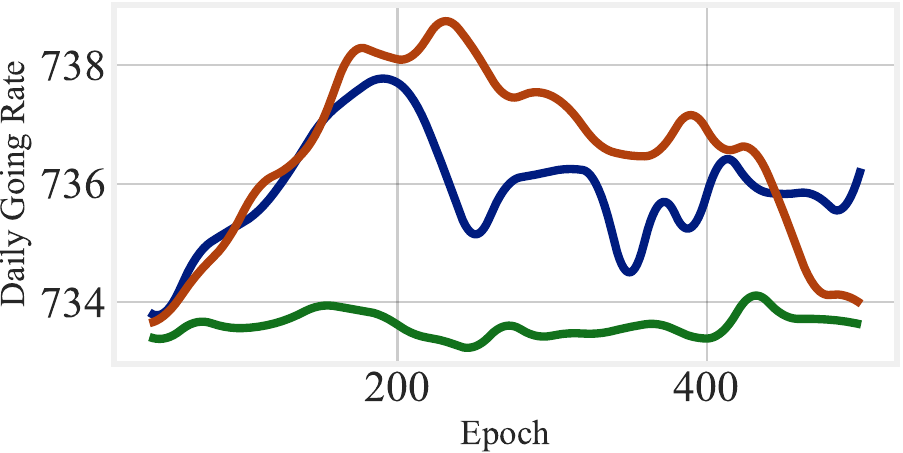}
  \end{subfigure}
  \hfill
  \begin{subfigure}[b]{0.32\textwidth}
    \includegraphics[width=\textwidth]{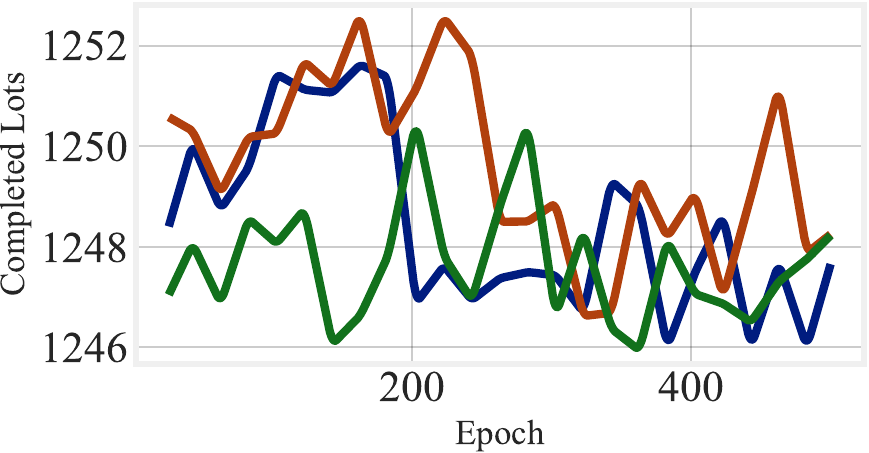}
  \end{subfigure}
  \hfill
  \begin{subfigure}[b]{0.32\textwidth}
    \includegraphics[width=\textwidth]{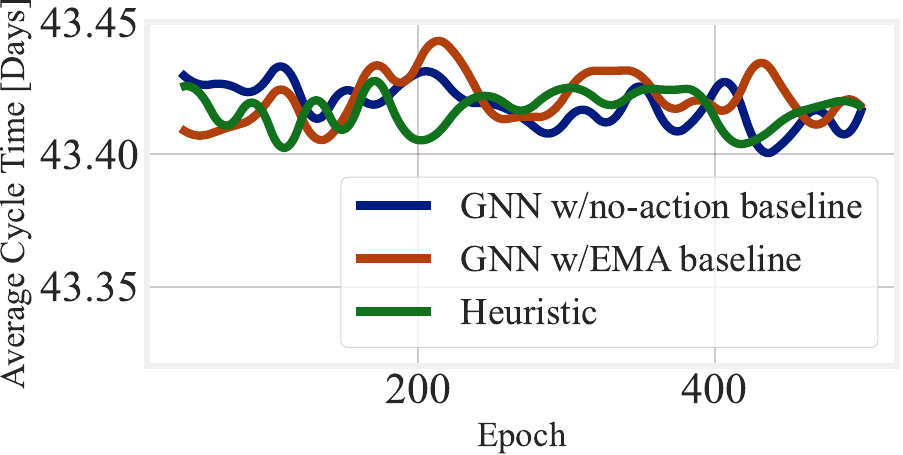}
  \end{subfigure}
  \caption{Fab-level KPIs in the five-week SMT2020 scenario.}
  \label{fig:smt2020_five_week_eval}
  \vspace{-0.25cm}
\end{figure}

To investigate potential causes for the observed decline after the initial improvement in the KPIs, we designed a restricted \textsc{SMT2020} scenario.
By limiting each epoch to 5 weeks and employing a constant seed for all simulator instances, we expect significantly lower variance in the rewards.
Furthermore, to constrain the action space, we omit dedication actions, allowing only for changes to the machines' efficiencies and uptimes.
In addition to our reward formulation using the difference in KPIs with and without actions, we include a variant using the difference between the current KPI and an exponential moving average (EMA) of the weekly KPIs from previous epochs.
The results shown in Figure~\ref{fig:smt2020_five_week_eval} suggest that the observed decline is not dependent on these factors.
As in the 25-week training, the learned actions trend towards their initial frequencies as the KPIs begin to decrease.
We are currently investigating appropriate regularization measures~\cite{dohare2023overcoming} in order to achieve reliable convergence to high-quality policies in long-term training runs.

\vspace{-0.15cm}
\section{Conclusion and Future Work}

We presented a GNN-based deep reinforcement learning framework for optimizing capacity planning decisions in semiconductor manufacturing.
Experiments using the small-scale \textsc{Minifab} testbed allowed us to select suitable hyperparameters and showed the general ability to learn effective capacity planning decisions.
We subsequently applied our approach to the larger-scale SMT2020 testbed, which includes many of the characteristics of real-world wafer fabs and approaches their scale, increasing throughput by about 1.8\%, which is a significant improvement in the semiconductor manufacturing context.

Our future work will explore three main avenues.
Firstly, we will explore improvements to the training stability to support gradual improvement of the policy during long-term training.
Secondly, to support real-world decision-making, the costs incurred by different actions must be taken into account and unprofitable actions avoided.
Finally, experiments on industry-scale simulation models will evaluate the scalability of our framework beyond common testbeds.
To support targeted proactive decisions in more dynamic settings, information about future wafer starts will be incorporated in the fab state representation.

\begin{credits}
\subsubsection{\ackname}
This research is supported 
in part by the National Research Foundation, Singapore under its AI 
Singapore Programme (AISG Award No: AISG3-RP-2022-031) and in part
 by the National Research Foundation, Singapore under its AI Singapore Programme (AISG Award No: AISG2-100E-2024-130).


\end{credits}

\bibliographystyle{splncs04}
\bibliography{main}

\end{document}